\documentclass[runningheads]{llncs}

\usepackage{eccv}

\usepackage{eccvabbrv}

\usepackage{graphicx}
\usepackage{booktabs}

\usepackage[accsupp]{axessibility}  

\usepackage{hyperref}

\usepackage{orcidlink}
\usepackage{multirow}

\begin{document}

\title{Text-to-Sticker: Style Tailoring Latent Diffusion Models for Human Expression} 

\titlerunning{Text-to-Sticker: Style Tailoring Latent Diffusion Models}

\author{Animesh Sinha$^*$,
Bo Sun$^*$,
Anmol Kalia$^*$,
Arantxa Casanova\thanks{Core contributors},\\
Elliot Blanchard,
David Yan,
Winnie Zhang,
Tony Nelli,
Jiahui Chen,
Hardik Shah,
Licheng Yu,
Mitesh Kumar Singh,
Ankit Ramchandani,
Maziar Sanjabi,
Sonal Gupta,
Amy Bearman\textsuperscript{\textdagger},
Dhruv Mahajan\textsuperscript{\textdagger}}
\authorrunning{Sinha et al.}

\institute{GenAI, Meta}

\maketitle

\begin{abstract}
  We introduce Style Tailoring, a recipe to finetune Latent Diffusion Models (LDMs) in a distinct domain with high visual quality, prompt alignment and scene diversity. We choose sticker image generation as the target domain, as the images significantly differ from photorealistic samples typically generated by large-scale LDMs. We start with a competent text-to-image model, like Emu, and show that relying on prompt engineering with a photorealistic model to generate stickers leads to poor prompt alignment and scene diversity. To overcome these drawbacks, we first finetune Emu on millions of sticker-like images collected using weak supervision to elicit diversity. Next, we curate Human-in-the-loop (HITL) Alignment and Style datasets from model generations, and finetune to improve prompt alignment and style alignment respectively. Sequential finetuning on these datasets poses a tradeoff between better style alignment and prompt alignment gains. To address this tradeoff, we propose a novel fine-tuning method called Style Tailoring, which jointly fits the content and style distribution and achieves best tradeoff. Evaluation results show our method improves visual quality by 14\%, prompt alignment by 16.2\% and scene diversity by 15.3\%, compared to prompt engineering the base Emu model for stickers generation.
\end{abstract}

\begin{figure}[ht!]
  \includegraphics[width=1.02\columnwidth]{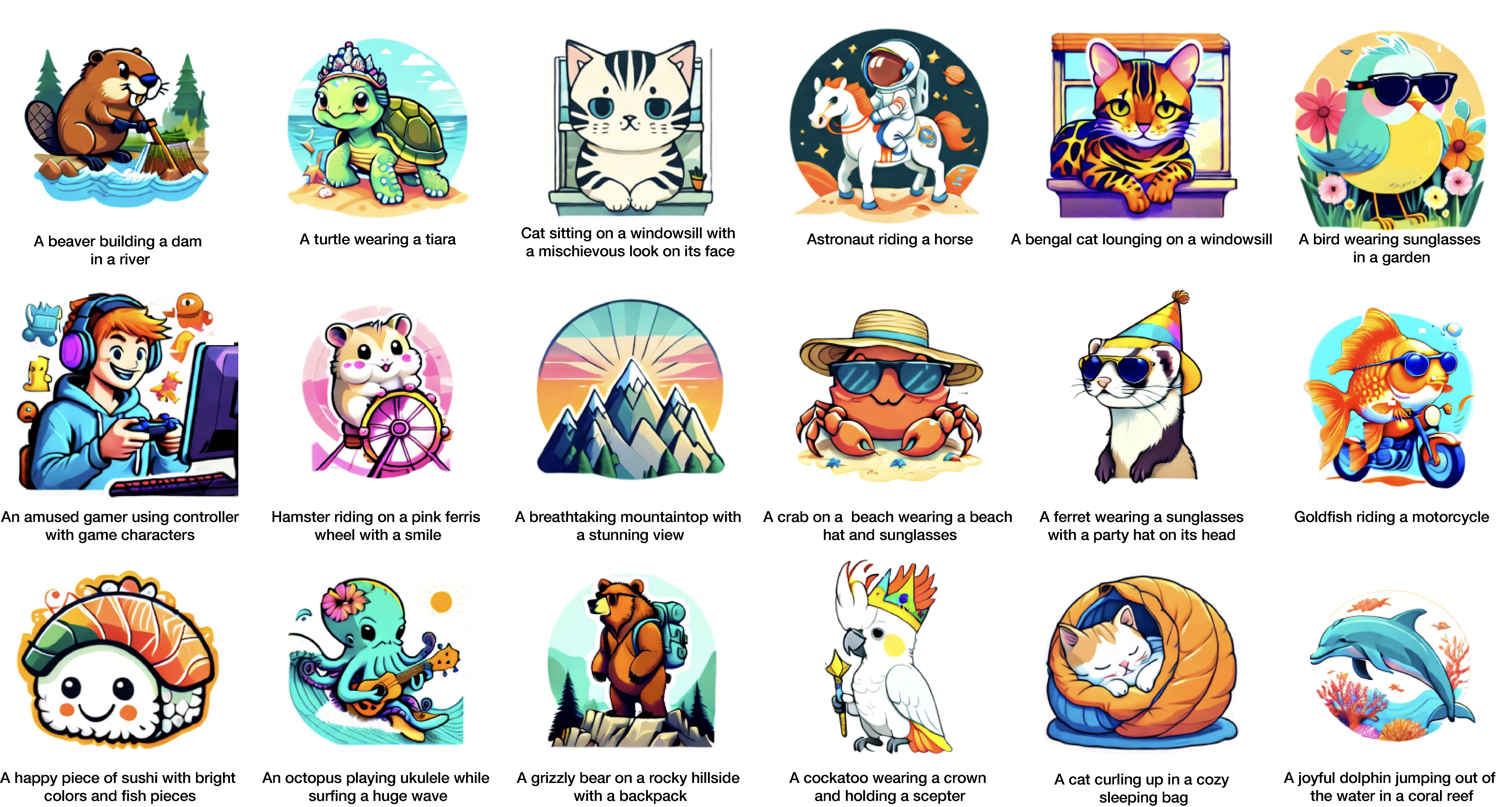}
  \vspace{-15pt}
  \caption{Stickers generated by our \textit{text-to-sticker} model. They are visually pleasing, diverse, and with high text faithfulness.}
    \vspace{-15pt}
\label{fig:teaser}
\end{figure}

\vspace{-5pt}
\section{Introduction}
\label{sec:intro}

There has been a substantial advancement in diffusion-based text-to-image models~\cite{Rombach_2022_CVPR, ramesh2022hierarchical, saharia2022photorealistic,Ruiz_2023_CVPR}, showcasing an unparalleled ability to understand natural language descriptions to generate high quality, visually pleasing images. These models empower users to conjure up entirely new scenes with unexplored compositions and generate striking images in numerous styles. Finetuning to a specific visual style has been explored in~\cite{gal2022textual, hu2022lora, sohn2023styledrop} as well as in the concurrent work~\cite{dai2023emu}, that finetunes Latent Diffusion Models (LDMs)~\cite{Rombach_2022_CVPR} to generate highly aesthetic images. 

Naively finetuning an LDM on a target style leads to a model whose distribution is aligned with the desired style, but comes at the cost of worse prompt alignment. We find that there exists a trade-off between consistently generating prompt aligned images and consistently generating on-style images. While current finetuning methods~\cite{gal2022textual, hu2022lora} have demonstrated the impressive capability of these models to produce highly aesthetic outcomes, they have not yet delved into mechanisms that simultaneously: (1) enhance prompt alignment, (2) improve visual diversity, (3) generate visually appealing images that (4) conform to a distinctive style. In this work, we are interested in training a model with all four aforementioned properties. In particular, we choose stickers generation as the motivating application for our proposed method.
 
We introduce a novel multi-stage fine-tuning approach aimed at optimizing both prompt alignment and visual diversity, while producing visually appealing stickers with a target style. Beginning with a domain alignment stage, weakly aligned sticker-like images are used to adapt the base text-to-image model \textit{Emu}~\cite{dai2023emu} to the sticker domain, followed by a human-in-the-Loop (HITL) stage to improve prompt alignment, and finally an experts-in-the-loop (EITL) stage to improve the sticker style aesthetics. Notably, in both HITL and EITL stages, the model is finetuned with \textit{generated} data only. HITL dataset consists of generated samples from the domain aligned model, chosen by human raters according to text faithfulness and quality guidelines. EITL style dataset contains generated images chosen by design experts using \textit{Emu} with prompt engineering. Finetuning the domain aligned model sequentially with HITL data and then style data leads to a tradeoff between style alignment on one hand, and prompt alignment and diversity on the other hand. Therefore, we propose a novel training method, \textit{Style Tailoring}, which combines and jointly optimizes for two data distribution in a single stage, and achieves the best tradeoff between prompt and style alignment. Style Tailoring decouples the LDM training objective into two parts: content and style loss. In the first few hundred denoising steps, the content loss is applied to ensure prompt alignment from content references, while the style loss is applied to the remainder of the timesteps to get the desired visual aesthetic. We also incorporate methods to achieve transparency and scene diversity in our pipeline to further enhance the visual please of generated stickers. We validate our approach by designing a robust human evaluation framework to measure visual quality, prompt alignment and scene diversity.

Our experiments show that the sequence in which fine-tuning steps are executed plays a crucial role in enhancing both visual quality and prompt alignment. We also show that the proposed recipe generalizes to more than one target style. Finally, the proposed methodology does not increase the latency with respect to the base, pre-trained LDM. We show generated images from our final model in Fig. \ref{fig:teaser} and quantitatively show improvements on visual quality, prompt alignment and scene diversity compared to prompt engineering \textit{Emu} in Table \ref{table:method_comparison}.

In summary, our main contributions are: 
\begin{enumerate}
    \item We propose a novel training method, called Style Tailoring, aimed at obtaining the best trade-off between prompt alignment, scene diversity and visual quality. 
    We show with qualitative examples that this method can generalize to other styles. 
    \item We conduct an extensive study of finetuning recipes to attain good performance along the axis of visual quality in a specific style domain,  prompt alignment and visual diversity. Through this study, we show the need of the domain alignment finetuning step, as well as the improvements brought by the HITL and Style datasets. 
    
    \item We propose a simple and effective solution to achieve transparency in LDM generations without introducing any additional latency.
    \item We propose a Prompt Enhancer module to enrich the scene diversity of the generated images, showing a novel use of an instruction tuned LLaMA model.
\end{enumerate}
\vspace{-5pt}
\section{Related Work}
\label{sec:related_work}
\vspace{-3pt}
\noindent\textbf{Text-to-Image Generation}. There has been a tremendous progress in the field of text-to-image generation in recent years. The use of the forward and reverse diffusion process~\cite{pmlr-v37-sohl-dickstein15} can achieve high fidelity in image generation~\cite{ramesh2022hierarchical, saharia2022photorealistic, dhariwal2021diffusion, balaji2023ediffi, Rombach_2022_CVPR, podell2023sdxl} compared to their GAN counterparts~\cite{zhang2017stackgan, kang2023scaling, salimans2016improved}. Among diffusion models, Latent Diffusion Models (LDMs)~\cite{Rombach_2022_CVPR} have demonstrated to be computation efficient and have found application in reconstructing images from human brain activity~\cite{Takagi_2023_CVPR}, video generation~\cite{Blattmann_2023_CVPR}, 3D environment generation~\cite{Kim_2023_CVPR}, image editing~\cite{brooks2022instructpix2pix}, controllable generation~\cite{zhang2023adding}, and much more. In this work, we focus on finetuning LDMs for a specific domain (stickers) and show their domain alignment capabilities.

\noindent\textbf{Human Preference Alignment}. 
Text-to-image diffusion models do not always generate images that are adequately aligned with the text description and human intent. To improve the alignment between text-to-image models and human preferences, \cite{lee2023aligning} proposes a reward-weighted likelihood maximization based on reward models trained from human feedback. \cite{wu2023better} demonstrates existing metrics~\cite{salimans2016improved,heusel2018gans,murray2012ava,pressman2023simulacra} for generative models have low correlation with human preferences. Then collects a dataset of human choices of generated images, and derives a Human Preference Score (HPS) for better alignment with human choices. \cite{xu2023imagereward} trains an ImageReward model using human choices that captures abstractions like aesthetic, body parts, and toxicity/biases. In our work, we leverage a human annotation pipeline to filter high-quality generated sticker images, and, we show that finetuning solely on high-quality generated data yields significant improvements in visual quality and prompt alignment, and attains a specific sticker style.

\noindent\textbf{Finetuning Text-to-Image Models}. Numerous finetuning strategies have been proposed in pursuit of high fidelity text-to-image generation. \cite{gal2022textual, hu2022lora} introduce new finetuning methods to align the pretrained diffusion models to a specific style, whereas,~\cite{chen2023photoverse, Ruiz_2023_CVPR, ruiz2023hyperdreambooth, avrahami2023breakascene} show high fidelity subject-driven generations using user provided images. 
\cite{sheynin2022knndiffusion} extends the conditioning of diffusion model to image embeddings retrieved by efficient k-nearest neighbors, enables generalizing to new distributions at test time by switching the retrieval database. \textit{Emu}~\cite{dai2023emu} shows that finetuning with few thousands of high-quality real images can significantly improve the visual quality of the generated images. 
\textit{Styledrop}~\cite{sohn2023styledrop} explores improving compositional power of text-to-image generation models, customizing content and style at the same time by adapter-guided sampling from adapters trained independently from content and style reference images. In our work, we show that there is a trade-off between style and text faithfulness during LDM finetuning. Then, we propose a novel finetuning approach called Style Tailoring, to balance such trade-off and optimize for both, without adding any modules or incurring extra latency at inference.
\vspace{-3pt}
\section{Model and Datasets}
\label{sec:method}

\vspace{-3pt}
\subsection{Text-to-Sticker model}\label{sec:model_overview}
\begin{figure*}[t!]
    \centering
    \includegraphics[width=0.95\linewidth]{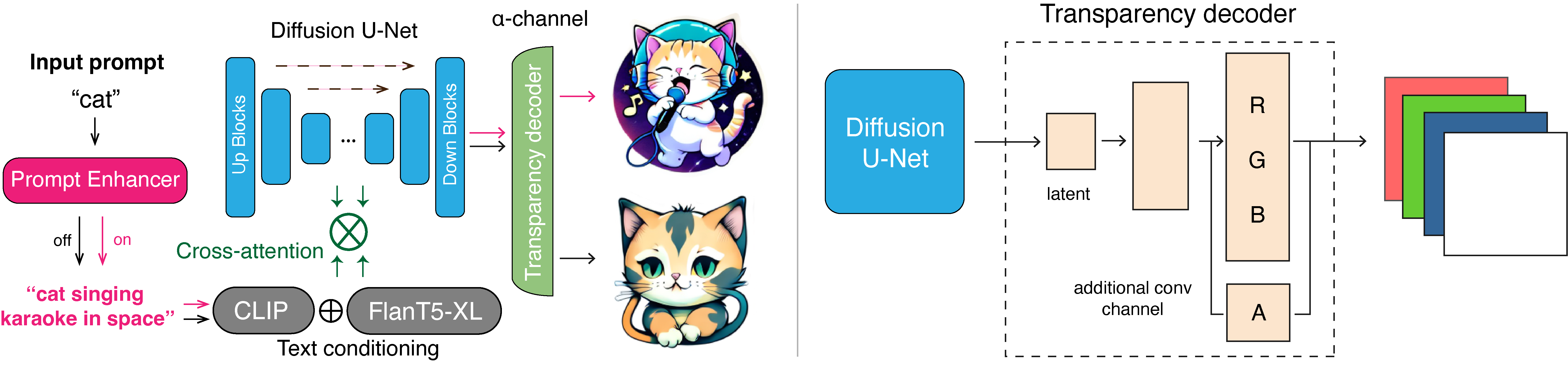}
    \vspace{-8pt}
    \caption{Architecture of our text-to-sticker model (left) and transparency decoder (right). The text-to-sticker model contains an optional prompt enhancer, a diffusion U-Net, and frozen text conditioning added via cross-attention. The transparency decoder is augemented from a regular VAE decoder, by adding one extra output channel at final output conv layer. The alpha-channel convolution weights are initialized with the average of R, G, B channels' weights.}
    \vspace{-10pt}
    \label{fig:model_arch}
\end{figure*}

Our text-to-sticker model (Fig. \ref{fig:model_arch}) consists of \textbf{(i)} Prompt Enhancer module, \textbf{(ii)} Text-guided Diffusion Module, and \textbf{(iii)} Transparency Module. Model output are sticker images with transparent background (alpha channel) conditionally generated on input or enhanced text prompts.

\noindent\textbf{Prompt Enhancer Module.}
Sometimes, user input prompts can be simple and abstract (e.g., ``love''). We create a Prompt Enhancer module to generate variations of input prompts, adding more descriptive details without altering its meaning. In favor of keeping our pipeline efficient, we decide to use the 1.4B instruction finetuned LLaMA model to re-phrase the input prompts in the Prompt Enhancer module. This model has the same architecture as Gopher 1.4B \cite{rae2021scaling} and is trained and instruction finetuned following \cite{touvron2023llama}. During inference, we prompt this LLaMA model with instructions (several examples of re-phrasing input prompts) and let it improvise another example for the input prompt. As an example, one random re-write of input prompt ``love'', is ``a wide-eyed puppy holding a heart''. With Prompt Enhancer module and instruction prompting, we manage to add a wide range of flavors and expressiveness without compromising the fidelity of user intentions. 

\label{sec:text2image_model}
\noindent\textbf{Text-guided Diffusion Module.} 
Our text-to-image module is a standard Latent Diffusion Model (LDM)~\cite{Rombach_2022_CVPR}, with a 2.6B trainable parameter U-net architecture~\cite{ronneberger2015u}, and initialized with the smallest version of the text-to-image model \textit{Emu}~\cite{dai2023emu} (\textit{Emu-256}), which generates images of size $256 \times 256$. As text conditioning, the concatenation of text embeddings from CLIP ViT-L~\cite{pmlr-v139-radford21a} and Flan T5-XL~\cite{JMLR:v21:20-074, chung2022scaling} are used. We use a 8-channel autoencoder in our model.

\noindent\textbf{Transparency Module.} Real stickers are rarely square, and transparent background usually makes stickers more visually pleasing. In order to balance visual style with other quality axes efficiently, we propose a novel solution for achieving transparency.  We mask the blank space around the generated sticker area with full transparency to create non-square stickers with transparent background. We achieve this by incrementing the output channel of the final convolution layer of the decoder
from 3 (RGB) to 4 (RGBA). The weights for the newly added alpha-channel are initialized as the mean of the weights for RGB channels, and all layers in the decoder 
are finetuned on the dataset discussed in Section \ref{sec:transparency_dataset}, while keeping the encoder frozen. Maintaining a frozen encoder allows for the replacement of the U-Net (e.g., trained for a different sticker style) without requiring retraining of the transparent decoder. This method of generating transparent images in text-to-image LDM model is novel, simple yet efficient. The additional computation is negligible since the only change is 3 to 4 channels in the final convolution layer.

\subsection{Datasets}\label{sec:datasets}
We utilize three separate datasets to train our model -- sticker \textbf{D}omain \textbf{A}lignment (DA) dataset, \textbf{H}uman-\textbf{I}n-\textbf{T}he-\textbf{L}oop (HITL) alignment dataset, and \textbf{E}xpert-\textbf{I}n-\textbf{T}he-\textbf{L}oop (EITL) style dataset. Images in the DA dataset are all real sticker-like images whereas the HITL and EITL datasets contain generated stickers only. Note that there's a trade-off between consistently generating prompt aligned and style aligned outputs. Hence, the need for two separate datasets that improve prompt alignment and style alignment respectively.
Additionally, we curate a dataset of stickers with transparency masks to train the transparency decoder.

\vspace{-2pt}
\subsubsection{Domain Alignment Dataset}\label{sec:pretraining_dataset}
We source 21M weakly aligned image-text pairs from a set of hashtags (\textsl{\#stickers}, \textsl{\#stickershop}, \textsl{\#cutestickers}, \textsl{\#cartoon}, \textit{etc.}) corresponding to sticker-like images, then apply two filtering steps. First, we filter out data with low image-text alignment calculated by CLIP score. Second, we apply an OCR model on the images and filter out images wherein detected OCR box $\ge8\%$ of the image area, to minimize text generated on stickers. Note that this dataset is collected primarily for visually aligning with sticker domain and has not been curated for high image-text alignment.

\vspace{-7pt}
\subsubsection{HITL Alignment Dataset}\label{sec:hitl_dataset}
The stickers domain dataset is noisy, and finetuning on this set alone is not sufficient to obtain high prompt alignment. 
To improve the model's prompt alignment, we systematically create prompt sets which cover relevant concepts for sticker generation, \textit{e.g.}, emotions, occupations, actions and activities, \textit{etc.} Then we generate stickers with the domain aligned model (Section \ref{sec:stickerspretraining}) and involve human annotators to filter for good quality images with high prompt alignment. We create three prompt buckets as described below:

\noindent\textbf{Emotion Expressiveness.} It contains human and animal emotions, consisting of 8 nouns which refer to humans (teen, kids, boy, girl, \textit{etc.}), 22 occupations (baker, doctor, lawyer, \textit{etc.}), and 83 animals. We perform Cartesian product between 36 common emotions and these human/animal concepts to form short phrases with correct grammar as prompts. For example, \textit{an angry hippo}, \textit{a sloth feeling tired}. 

\noindent\textbf{Object Composition.} It contains prompts composed by the Cartesian product of aforementioned human/animal concepts with ``single-action'' and ``pair-action''. Here ``single-action'' is defined as an action that can be performed by a single object, \textit{e.g.} \textit{a bear drinking coffee} or \textit{a dog playing frisbee}. And ``pair-action'' is defined as actions that involves two subjects, \textit{e.g.} \textit{a turtle giving present to a rabbit} or \textit{a cat playing with a giraffe}.

\noindent\textbf{Scene Diversity.} We leverage the instruction finetuned 1.4B LLaMA model to collect prompts that are hard to be structurally composed by sentence templates, like ``landscape'' (\textit{e.g.}, \textit{river flows down the valley}), and ``activities'' (\textit{e.g.}, \textit{family trip}). To be noted, the LLaMA model here is the same as in the Prompt Enhancer (Section \ref{sec:model_overview}) but the instruction prompting is different. In Prompt Enhancer, LLaMA model re-writes a given input prompt, but here the prompts are composed from scratch. 

For the Emotion, Scene Diversity and Object Composition sets we generate 5, 5 and 6 images per prompt, respectively.  Human annotators rate the generated stickers as pass/fail based on guidelines for visual quality (particularly for faces and body parts) and prompt alignment. The stickers labeled as \textit{pass} become our HITL alignment dataset. Details on the pass-rate, number of training images in the HITL alignment dataset and visual examples from each bucket are shown in the supplementary.


\begin{figure*}[t]
    \centering
    \includegraphics[width=0.98\linewidth]{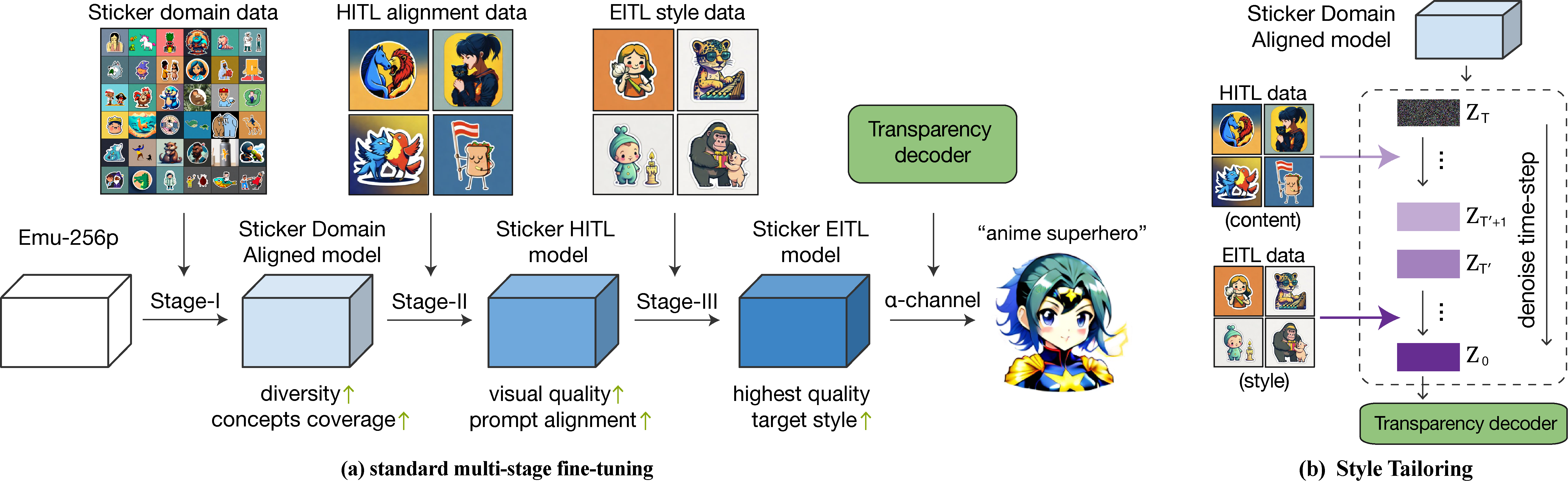}
    \vspace{-5pt}
    \caption{Illustration of our \textit{text-to-sticker} model finetuning recipe. \textbf{(a)} Standard multi-stage fine-tuning. \textbf{(b)} Our proposed method, Style Tailoring. In Style Tailoring, we implement a phased dataloader such that the U-Net denoising steps $T$ to $T'+1$ are trained with HITL alignment data (content distribution $p_{content}$), and denoising steps $T'$ to $0$ are trained with EITL data (style distribution $p_{style}$).}
    \vspace{-4pt}
    \label{fig:ft_recipe}
\end{figure*}


\vspace{-3pt}
\subsubsection{EITL Style Dataset}\label{sec:style_dataset}
\vspace{-2pt}
Besides general visual quality and prompt alignment, we also want to obtain a \textit{text-to-sticker} model that adhere to a target sticker style criteria (color, sharpness, linework, layout, shading, \textit{etc.}). While non-expert human raters perform well on the task of judging prompt alignment and visual quality, their label quality for the style criteria are quite low. Instead, we find that design experts are much more reliable in selecting generated stickers with target style. To collect the style dataset, we generate stickers using the \textit{Emu-256} model with prompt engineering. We choose \textit{Emu-256} for this because we find that, with prompt engineering carefully designed by experts, it has the best ability to generate images in the desired style. However, since the \textit{Emu-256} model has low prompt alignment as illustrated in Table \ref{table:method_comparison}, we're only able to collect data from this model for single subject prompts and not for composition prompts.  Our final EITL style dataset contains 4235 stickers hand curated by design experts, with a few random examples shown in the supplementary.

\vspace{-3pt}
\subsubsection{Transparency Dataset}\label{sec:transparency_dataset}
\vspace{-1pt}
We curate a dataset of images with transparency masks to train the Transparency Module (Section \ref{sec:model_overview}). First, we use Segment Anything Model~\cite{kirillov2023segment} to generate foreground masks on a subset of 200K stickers from our domain alignment dataset.
Then, we refine these masks with a human curation process, that is accelerated given that the annotators do not need to start segmenting from scratch. 

\subsection{Multi-stage Fine Tuning}\label{sec:finetuning}

In this section, we describe the steps in our multi-stage fine-tuning recipe which turns the general purpose text-to-image model into a specialized text-to-sticker model. Starting with (i) domain alignment finetuning, followed by (ii) prompt alignment on HITL data and (iii) style alignment on EITL style data. We find there is a clear tradeoff between prompt and style alignment, and propose a novel finetuning method Style Tailoring -- the best in-between solution maintaining both prompt and style alignment.

\vspace{1pt}
\noindent\textbf{Training Objectives.} In all alignment stages, we empirically observe finetuning the full U-Net yields the best results. The U-Net parameters $\epsilon_\theta$ are updated by optimizing the noise reconstruction objective in all three finetuning stages $\mathcal{D}$ $\in$ \{Domain Alignment, HITL, EITL\} dataset, 
\begin{equation*}
    \mathcal{L}(\theta;\epsilon,t) = \mathop{\mathbb{E}}_{\epsilon \sim \mathcal{N}(0,1)} {\scriptstyle (x, y)\sim\mathcal{D}} \bigl( \Vert \epsilon - \epsilon_\theta(\mathcal{E}(x), \mathcal{T}(y); t)  \Vert^2\bigr)
\end{equation*}
where $\epsilon$ denotes the Gaussian noise sample, $(x,y)$ denotes the image-text pair, $\mathcal{E}$ denotes the image autoencoder, $\mathcal{T}$ denotes text encoder and $t$ denotes the denoising timesteps.

\vspace{-5pt}
\subsubsection{Domain Alignment.}\label{sec:stickerspretraining}
Relying on prompt engineering to generate stickers with the general text-to-image model (\textit{Emu-256}) leads to poor prompt alignment and low scene diversity (details explained in Section \ref{subsec:aba_baseline}). One reason this happens is the \textit{Emu} models have been finetuned on a small high quality dataset. To spur on diverse sticker generations, we first align \textit{Emu-256} closer to the sticker domain by finetuning with our Domain Alignment (DA) dataset (Section \ref{sec:pretraining_dataset}), which contains 21M sticker image-text pairs. DA dataset contains diverse stickers in assorted styles with loosely aligned captions, we find the domain alignment finetuning largely improves diversity and weakly improves prompt alignment, improvements are quantified in Table \ref{table:method_comparison}.

\vspace{-4pt}
\subsubsection{Prompt Alignment and Style Alignment.} \label{sec:alignmentfinetuning}
To further improve prompt and style alignment, we finetune the domain aligned model with the HITL alignment dataset (Section \ref{sec:hitl_dataset}) and the EITL style dataset (Section \ref{sec:style_dataset}). The former has high prompt alignment, the latter contains hand-curated stickers with target style. In our standard finetuning recipe (Fig. \ref{fig:ft_recipe}a), we first finetune the domain aligned checkpoint on HITL dataset for better prompt alignment, and then we bake-in the target style by fine-tuning the HITL checkpoint on EITL style dataset. We notice a clear tradeoff between prompt alignment and style alignment. While finetuning on EITL style dataset hugely improves style alignment, it erases some of the prompt alignment gains from HITL. This motivates us to develop the novel finetuning method called Style Tailoring, which achieves the best balance between the two objectives, without adding any extra modules or latency.

\vspace{-3pt}
\subsubsection{Style Tailoring.}
\vspace{-5pt}


In the standard LDM training, the timestep $t \sim [0, T]$ is uniformly sampled. Our key observation is that when denoising the later timestamps that are closer to the noise sample $z_T$, the model learns to generate the coarser semantics -- the \textit{content} of the image. And when denoising the earlier timestamps that are closer to the denoised image latent $z_0$, the model learns the fine-grained details -- the \textit{style} of the image.

Different from standard LDM training which denoises latents for decoding images from a single training data distribution $p_{data}$, we propose to train it to denoise latents from two distributions conditioned on timesteps (Fig. \ref{fig:ft_recipe}b). Given a sampled timestep $t$, we train the denoising U-Net with data points sampled from a content distribution $p_{content}$ for timestamps $t$ closer to noise $t\in[T, T')$, and data points sampled from a style distribution $p_{style}$ for timestamps closer to the final image latent. In our case, HITL alignment dataset $D_{hitl}$ represents the content distribution $p_{content}$, and EITL style dataset $D_{style}$ represents the style distribution $p_{style}$. Formally, $\forall \epsilon\in\mathcal{N}(0,1)$, the joint objective can be written as
\vspace{-1pt}
\begin{equation*}
\begin{split}
    \mathcal{L}(\theta;\epsilon,t) & = \mathcal{L}_{content}(\theta;\epsilon,t) + \mathcal{L}_{style}(\theta;\epsilon,t) \\ & = \mathop{\mathbb{E}}_{t\in(T',T]} {\scriptstyle (x, y)\sim\mathcal{D}_{hitl}} \bigl( \Vert \epsilon - \epsilon_\theta(\mathcal{E}(x), \mathcal{T}(y); t)  \Vert^2\bigr) \\
                                   & + \mathop{\mathbb{E}}_{t\in[0_,T']} {\scriptstyle (x, y)\sim\mathcal{D}_{style}} \bigl( \Vert \epsilon - \epsilon_\theta(\mathcal{E}(x), \mathcal{T}(y); t)  \Vert^2\bigr)
\end{split}
\end{equation*}
The timestep $T'$ represents the timestep cutoff for using the content and style datasets. Experiments in Section \ref{sec:sec5} show that Style Tailoring offers a superior middle ground, with strong prompt alignment while also generating images that aligns well with the target style. 

\vspace{-3pt}
\subsection{Training Details}
\label{subsec:tr_details}
\textbf{Domain Alignment.} We train the model with global batch size 2,240 on $D_{da}$ dataset for 300K steps, using learning rate 1e-5 with linear warm up followed by a constant schedule. It takes around 19,200 A100 gpu hours for stickers domain alignment. We use \textit{eps} parameterization to train the model instead of \textit{v}~\cite{salimans2022progressive} as training using \textit{eps} parameterization led to better body shapes and quality. 

\vspace{3pt}
\noindent\textbf{Prompt Alignment and Style Alignment.} For all subsequent finetuning steps, we use a lower learning rate of 5e-6 and a global batch size of 256. We initialize from the domain aligned model and finetune for 8k steps on $D_{hitl}$ for prompt alignment. Once trained, we further fine-tune this model for 3k steps on style reference $D_{style}$. We  stop early at 3k steps since we observe that we get best results during the warm-up period with less over-fitting.

\vspace{3pt}
\noindent\textbf{Style Tailoring.} In Style Tailoring, we train the model for 5k steps. We empirically set $T'$=$900$, which means the 100 timestamps closer to sampled noise are trained with $D_{hitl}$, and the remaining 900 timestamps are trained with $D_{style}$. In each batch, training data points from $D_{hitl}$ and $D_{style}$ are sampled in a 1:1 ratio. We show the effect of $T'$ on the metrics in Table \ref{table:st_threshold}.
\vspace{-3pt}
\section{Evaluation Dataset and Metrics}\label{sec:sec4}

We use a combination of human evaluations and automatic evaluation metrics to understand the performance of the models regarding the (i) visual quality (ii) prompt alignment (iii) style alignment and (iv) scene diversity, of sticker generations.

\noindent\textbf{Evaluation Dataset}. For (i) sticker visual quality, we curated a list of 750 prompts -- daily activities, aspirational phrases, object compositions, \textit{etc} --, and generated two images per prompt.
For (ii) prompt alignment, we curated 300 hard compositional prompts -- 100 for emotion expressiveness and 200 for actions and interactions. In this case, ten images are generated for each prompt. 
Same seed and starting noise are used when generating stickers for different models, to ensure accurate and fair comparisons. For (iii) style alignment and (iv) scene diversity, we prepare a style reference dataset containing around 4150 images. The style reference data is collected by the same design experts and same procedure as in Section \ref{sec:style_dataset}, but held-out as a test set. To measure style alignment and scene diversity, we generate one and two images per prompt respectively.

\noindent{\textbf{Human Evaluation}.} We design comprehensive human annotation tasks to measure model performance on evaluation dataset. For (i) visual quality, we present annotators with a sticker and ask them to assess whether it meets the guidelines based on nine different criteria -- \textsl{Color}, \textsl{Sharpness}, \textsl{Linework}, \textsl{Detail}, \textsl{Lighting}, \textsl{Centering and Leveling}, \textsl{Flat 2D}, \textsl{Human Faces}, and \textsl{No Text}. We collaborate with design experts when designing guideline rubric for each visual axes. For (ii) prompt alignment, we present raters with a text-sticker pair and ask them to evaluate whether the sticker accurately passes five key aspects -- \textsl{Subject}, \textsl{Quantity}, \textsl{Face \& Emotion}, \textsl{Action}, and \textsl{Body Parts}. For each annotation job, we use three multi-reviews and take their majority vote as the final label. 

\noindent{\textbf{Automatic Evaluation Metrics.}} To measure (iii) style alignment, we propose Fr\'echet DINO Distance (FDD), with DINOv2~\cite{oquab2023dinov2} as a feature extractor instead of the conventionally used InceptionV3~\cite{Szegedy_2016_CVPR}. InceptionV3 is trained on ImageNet~\cite{deng2009imagenet},
and it performs poorly when generalizing to other out-of-distribution domains, such as stickers. Instead, DINOv2 is a self-supervised method that has been shown to generalize better. To measure (iv) scene diversity, we use LPIPS~\cite{zhang2018unreasonable}, as is a standard practice in the conditional image generation community~\cite{Zhao_2019_CVPR,casanova2021instanceconditioned}. Higher LPIPS indicates higher scene diversity (perceptual similarity) between two generated images using the same conditioning.

\section{Experiments} \label{sec:sec5}

Our goal is to train a model which generates visually appealing stickers and are faithful to the text prompt while being in the target visual style. In this section, we show experiments on model baseline, analysis of each finetuning stage, results and generalization of style tailoring. Finally, we discuss the effect of LLaMA for prompt expansion and share quantitative metrics for our transparency decoder.

\vspace{-3pt}
\subsection{Baseline}\label{subsec:aba_baseline}
We consider applying sticker-style prompt engineering (PE) on general purpose text-to-image model as our baseline, PE word choices are conjugated by design experts to achieve the desired style. Compared to Stable Diffusion v1 (SDv1-512)~\cite{Blattmann_2023_CVPR}, \textit{Emu-256} has a higher success rate of generating desired sticker style with good quality, we therefore use \textit{Emu-256} as our baseline and the foundation model for text-to-sticker. We observe two limitations on this \textit{Emu-256} + PE baseline -- (i) poor prompt alignment (76\% pass-rate) and (ii) low scene diversity (0.469 LPIPS)(Table \ref{table:method_comparison}). The baseline model always generates similar looking subjects and postures, and fails on compositions for common concepts. Visual examples demonstrating the limitations of this baseline model are shown in the supplementary. Collecting HITL data directly from this baseline results in a low diversity dataset and finetuning with it further reduces diversity. We finetune the baseline model on Domain Alignment dataset first to improve diversity. 

\begin{figure*}[t!]
    \centering
    \includegraphics[width=0.96\linewidth]{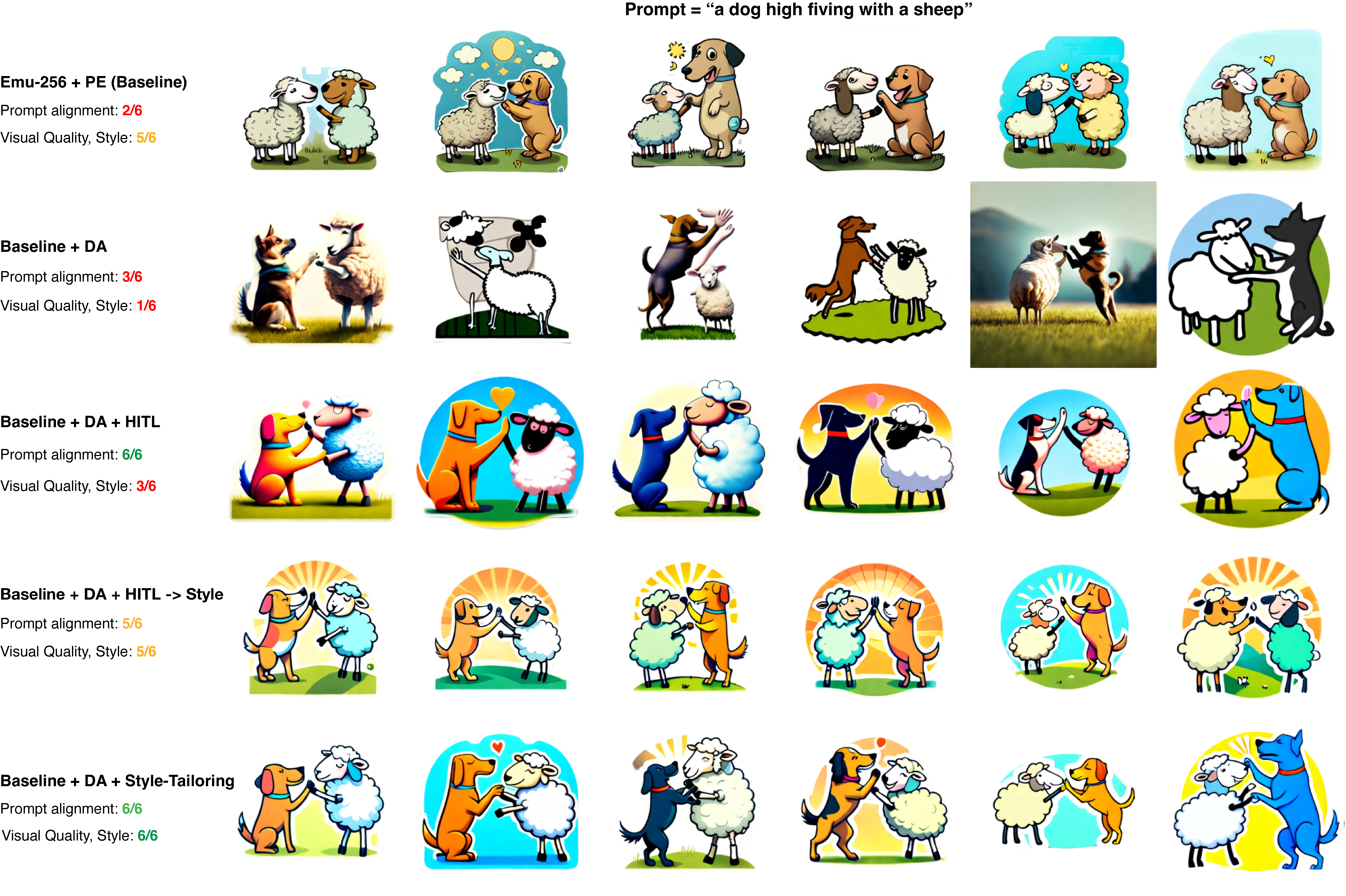}
    \vspace{-5pt}
    \caption{Qualitative results for the models in Table \ref{table:method_comparison}. Baseline (Row 1) lacks prompt alignment and diversity, domain aligned model (Row 2) improves alignment and diversity but is much worse in quality. Multi-stage finetuning (Rows 3 \& 4) face a trade off between prompt and style alignment. Style Tailoring (Row 5) offers the best results in both prompt and style alignment. More qualitative examples in the supplementary.}
    \vspace{-4pt}
    \label{fig:main_ablation_figure}
\end{figure*}

\begin{table*}[ht]\small
\caption{Evaluation results for all models and finetuning recipes. Target Style and Scene Diversity are measured by automatic metrics FDD and LPIPS respectively. Visual Quality and Prompt Alignment are measured by human annotation with multi-review = 3. Best results are shown in bold numbers, best trade-offs are underlined. The Visual Quality human eval is omitted for R2, R3 \& R4 as they either deviated too much from the target visual style or prompt alignment. Best trade-off: Style-Tailoring.}
\renewcommand{\tabcolsep}{6pt}
\centering
\resizebox{\linewidth}{!}{
 \begin{tabular}{llcccc}
\toprule
 & \textbf{Model} & $\downarrow$\textbf{FDD} &  $\uparrow$\textbf{LPIPS} &  $\uparrow$\textbf{Quality} (\%) & $\uparrow$\textbf{Prompt Alignment} (\%) \\ 
 \midrule
R0 & SDv1-512 + PE & 776.0 & 0.483 & 44.8& 30.9 \\
R1 & Emu-256 + PE (Baseline) & \textbf{168.30} $\pm$ 1.20 & 0.469  $\pm$ 0.005 & 65.2 & 76 \\ 
\midrule
R2 & Baseline + DA &  796.82 $\pm$ 5.55 & \textbf{0.696} $\pm$ 0.002 & - & 82.4\\ 
R3 & Baseline + DA + HITL & 374.29 $\pm$ 1.54 & 0.570 $\pm$ 0.006 & - & \textbf{91.1}\\ 
R4 & Baseline + DA + Style & 264.81$\pm$ 0.83 &  0.458 $\pm$ 0.006 & - & 74.3\\ 
R5 & Baseline + DA + Style$\,\to\,$HITL & 457.05 $\pm$ 0.61 & 0.397 $\pm$ 0.007 & 64.9 & 79.8\\ 
R6 & Baseline + DA + HITL$\,\to\,$Style & 301.10 $\pm$ 2.48 & 0.466  $\pm$ 0.006 & \textbf{75.1} & 85.3 \\ 
\midrule
R7 & Baseline + DA + Style-Tailoring & \underline{290.95} $\pm$ 2.37 & \underline{0.541}  $\pm$ 0.001 & \underline{74.3} & \underline{88.3}\\ 
\bottomrule
\end{tabular}}
\vspace{-15pt}
\label{table:method_comparison}
\end{table*}

\vspace{-3pt}
\subsection{Analysis of Multi-stage Finetuning}\label{subsec:aba_finetuning}
\noindent{\textbf{Effectiveness of Domain Alignment}}. Table \ref{table:method_comparison}, Row 2 vs 1 (R2 vs R1) shows that Domain Alignment substantially increases scene diversity (LPIPS 0.469 $\rightarrow$ 0.696) and moderately increases prompt alignment (76\% $\rightarrow$ 82.4\%) as well. This meets with our expectation since the DA dataset contains weakly-aligned text-sticker pairs from multiple styles. However, the sticker domain aligned model moves away from the target style (FDD 168.30 $\rightarrow$ 796.82), since the DA dataset contains stickers in mixed quality and style. We therefore introduce the subsequent HITL alignment and EITL style finetuning to boost prompt alignment and bring back the target style. Due to the improved prompt alignment of this model, we achieve a higher pass-rate when utilizing the domain-aligned model for collecting HITL alignment data. As a result, we can obtain the same amount of data with fewer annotators or in less time, leading to cost and resource savings. 

\noindent{\textbf{Effect of HITL Alignment Finetuning.}} Table \ref{table:method_comparison}, R3 vs R2 shows that finetuning the domain aligned model with HITL dataset largely improves prompt alignment (82.4\% $\rightarrow$ 91.1\%). Besides, the model moves closer to the desired style (FDD 796.82 $\rightarrow$ 374.29), given annotations guidelines contain criteria for visual quality. Fig. \ref{fig:main_ablation_figure} qualitatively shows the HITL model (3rd row) has much better prompt alignment than baseline (1st row) and domain aligned model (2nd row).


\noindent{\textbf{Effect of EITL Style Finetuning.}} Table \ref{table:method_comparison}, R4 vs R2 and R6 vs R3 show that finetuning the model with EITL style dataset further improves the target style alignment (FDD 796.82 $\rightarrow$ 264.81 and 374.29 $\rightarrow$ 301.10). This is because the design experts have higher accuracy labeling according to the style criteria. However, we notice the prompt alignment and scene diversity reduce when finetuning with the style dataset in both cases. 


\noindent{\textbf{Effect of HITL and EITL Finetuning Order.}} For this ablation, we perform standard finetuning in two steps and experiment with the order of finetuning: (a) we use the Baseline+DA model and collect the HITL dataset, finetune on it and then finetune on the Style dataset. We name this order as \textit{HITL\textrightarrow Style}; we test the reverse order, where (b) we finetune the stickers Baseline+DA on the Style dataset, then use the resulting model to collect HITL data and further finetune the model on it. We name this order as \textit{Style\textrightarrow HITL}. In Table \ref{table:method_comparison}, R5 and R6 shows that R6 is superior across all metrics, showing that the best order is first finetuning on HITL data and finally on Style data.

Overall, we observe that keys to human-in-the-loop finetuning are (i) having a good-enough and diverse foundation model to apply HITL on and, (ii) applying Expert-in-the-loop (EITL) on top of a stronger HITL model, to really let the style finetuning shine. It's worth mentioning that conducting HITL fine-tuning at an earlier stage offers the advantage of removing the need to collect HITL data again each time the target style changes. 


\vspace{-3pt}
\subsection{Style Tailoring: Best Trade-off}\label{subsec:st}
Comparing with sequential finetuning (R6) in Table \ref{table:method_comparison}, style-tailored model (R7) improves prompt alignment by $+3.5\%$, scene diversity by $+16.2\%$ (LPIPS 0.466 $\rightarrow$ 0.541 ), with superior style alignment (FDD 301.10 $\rightarrow$ 290.95, $+3.8\%$) and similar visual quality (75.1\%$\rightarrow$ 74.3\%, $-0.8\%$). Style Tailoring offers the best trade-off between all metrics of consideration -- prompt alignment, quality, diversity and style. While different models have the best performance in a single metric, they all come with significant degradation in other metrics. It is expected that the baseline \textit{Emu-256} (R1) has the best style alignment, because the style reference test set is curated from it. Overall, the style-tailored model obtains second-best results from all perspectives, with close-to-best performance. 

\vspace{-10pt}
\begin{table}[h]\small
\caption{Effect of $T'$ in the Style Tailoring method.}
\vspace{-5pt}
\renewcommand{\tabcolsep}{4pt}
\centering
 \begin{tabular}{lcccc}
\toprule
 & \textbf{T'} & $\downarrow$\textbf{FDD} &  $\uparrow$\textbf{LPIPS} &  $\uparrow$\textbf{Prompt Alignment}\\ 
 \midrule
& 600 & 260.25 & 0.567 & 0.75\\
& 700 & 235.59 & 0.569 & 0.75\\ 
& 800 & 214.48 & 0.545 & 0.76\\ 
& 900 & 204.48 & 0.517 & 0.76\\ 
& 950 & 200.91 & 0.502 & 0.70\\ 
\bottomrule
\end{tabular}
\label{table:st_threshold}
\end{table}

\vspace{-15pt}
\noindent\textbf{Effect of $T'$ in Style Tailoring}. Table \ref{table:st_threshold} shows that larger values of $T'$ result in better style alignment (lower FDD) but lower diversity (lower LPIPS). The prompt alignment has similar values throughout except it drops 5\% when $T'=950$, which means that 50 timesteps are not enough to get prompt alignment benefits from HITL. Instead, training for more than 100 timesteps with this data yields similar performance. This confirms the hypothesis that the \textit{content}, related to prompt alignment, is mostly formed in earlier timesteps, instead of later ones. Therefore, $T'=900$ gets the best trade-off between both datasets.

\begin{figure}[t]
    \centering
    \includegraphics[width=0.8\linewidth]{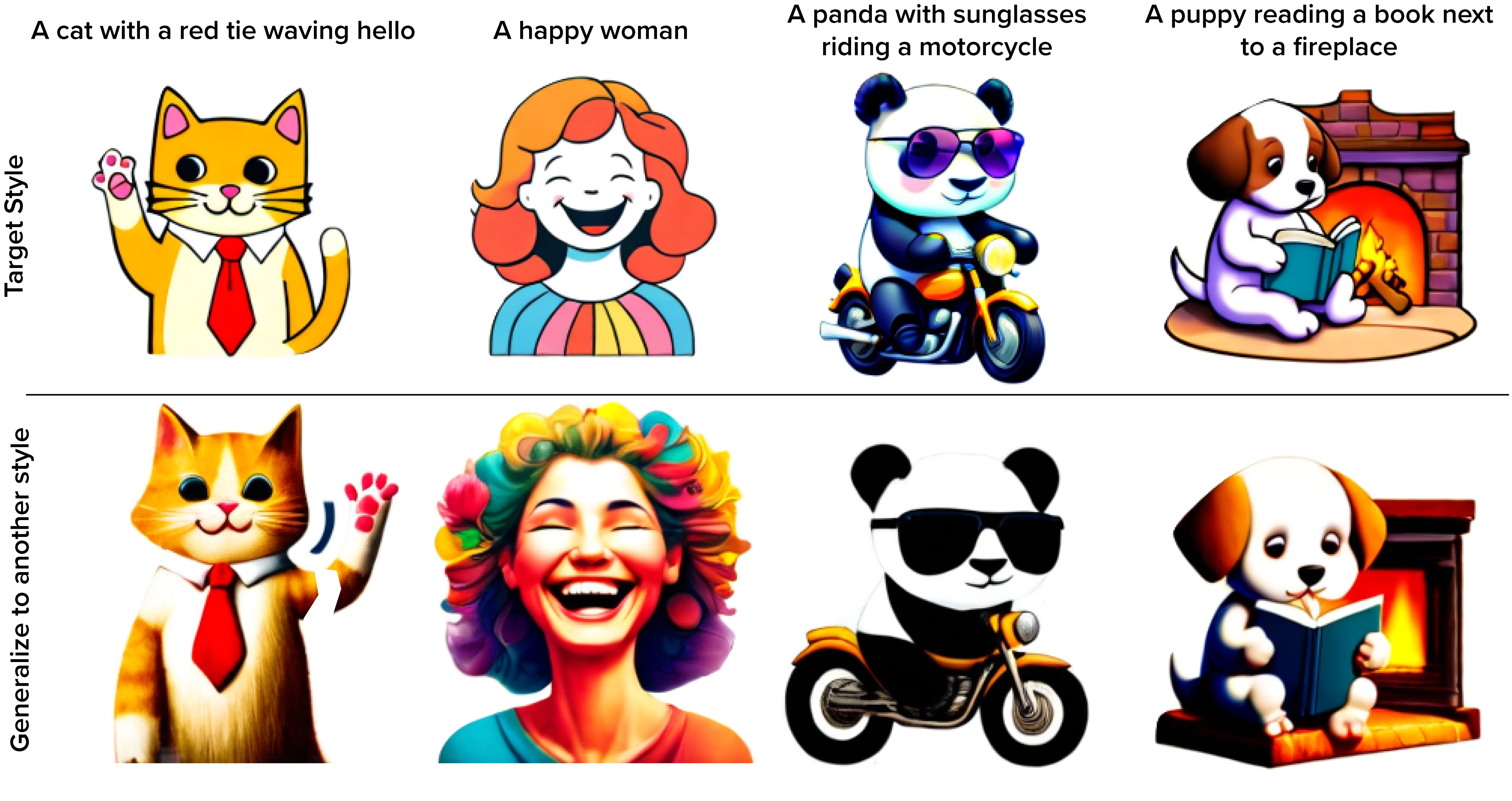}
    \vspace{-0.2in}
    \caption{Generalization of Style tailoring to multiple styles: our final, target graphic style (top row) and alternate volumetric style (bottom row).}
    \vspace{-6pt}
    \label{fig:st_style_comparison}
\end{figure}

\noindent\textbf{Generalization of Style Tailoring}.  We created another dataset with a distinct graphic style -- 3D looking stickers, contrary to the 2D look in the original target style -- and conducted experiments to determine if it could be applied to other styles. As depicted in Figure \ref{fig:st_style_comparison}, the method proved effective for this alternative style, demonstrating that because it doesn't rely on any style-specific assumptions, it can indeed generalize across various styles.


\vspace{-3pt}
\subsection{Effect of LLaMA for Prompt Expansion}
We ablate the effect of Prompt Enhancer module on scene diversity and prompt alignment. In human evaluation of scene diversity, we find that incorporating LLaMA into the pipeline results in a win rate of 67\%, a tie rate of 14\%, and a loss rate of 19\% when compared to cases where it is not included. The automatic scene diversity metric LPIPS also increases from 0.541 to 0.61 (+12.8\%). Human evaluations also show that the scene diversity is improved without losing text faithfulness, prompt alignment metrics stay the same (+/- 0.1\%). 

\subsection{Evaluating the transparency decoder}
We ran human eval on our quality eval set (750 prompts x 2 images) and found that 49.6\% stickers had perfect transparency mask, while 38.5\% had tiny dots or holes in the alpha channel. Only 11.9\% had no transparency mask.  

\vspace{-3pt}
\section{Limitations}
\vspace{-3pt}
\noindent\textbf{Foundational Text-to-Image Model Limitations}. Despite multiple rounds of fine-tuning in the stickers domain, the model occasionally generates photorealistic images when prompted with extremely rare concepts not seen during pretraining and finetuning. This can be minimized by continuously enriching the training data with new concepts over time. 

\noindent\textbf{Human Evaluation Limitations}. We built a robust evaluation framework with clear guidelines, but there could be subjectivity when benchmarking generative models due to human preferences. We have tried to construct a prompt set and balanced image reference datasets inline with the current trends. However, it's important to note that these results may evolve over time as trends shift.

\vspace{-3pt}
\section{Conclusion}
In our study, we illustrate the concurrent fine-tuning of diffusion models for both prompt alignment and visual quality within the domain of stickers. Our primary focus in this research centers around the idea that thoughtfully chosen generated images, in conjunction with our proposed approach, Style Tailoring, can result in visually pleasing and highly-aligned generations. We also discuss the tradeoffs of applying prompt engineering on powerful base models to achieve a desired style. Furthermore, we establish the generalizability of our method across multiple sticker styles, and prove its effectiveness through detailed human evaluation tasks.

\section*{Acknowledgements}
We thank Tamara Berg, Emily Luo, Sweta Karlekar, Luxin Zhang, Nader Hamekasi, John Nguyen, Yipin Zhou, Matt Butler, Logan Kerr, Xiaoliang Dai, Ji Hou, Jialiang Wang, Peizhao Zhang, Simran Motwani, Eric Alamillo, Ajay Menon, Lawrence Chen, Vladan Petrovic, Sean Dougherty, Vijai Mohan, Ali Thabet, Yinan Zhao, Artsiom Sanakoyeu, Edgar Schoenfeld, Jonas Kohler, Albert Pumarola, Ankit Jain, Shuming Hu, Li Chen, May Zhou, Sean Chang Culatana, Harihar Subramanyam, Bonnie Zhou, Jianfa Chen, Emily Shen, Uriel Singer, Shelly Sheynin, Vincent Cheung, Devi Parikh, Tali Zvi, Peter Vajda, Roshan Sumbaly, Manohar Paluri, Ahmad Al-Dahle and others who supported, contributed and provided feedback on the work throughout.


%
%
\bibliographystyle{splncs04}
\bibliography{main}

\clearpage
\setcounter{linenumber}{1}
\appendix
\counterwithin{equation}{section}

\section*{Appendix}

\section{Visualizations}
\label{sec:visualizations}
\subsection{HITL Alignment Dataset}
\begin{figure}[ht]
    \centering
    \includegraphics[width=1\linewidth]{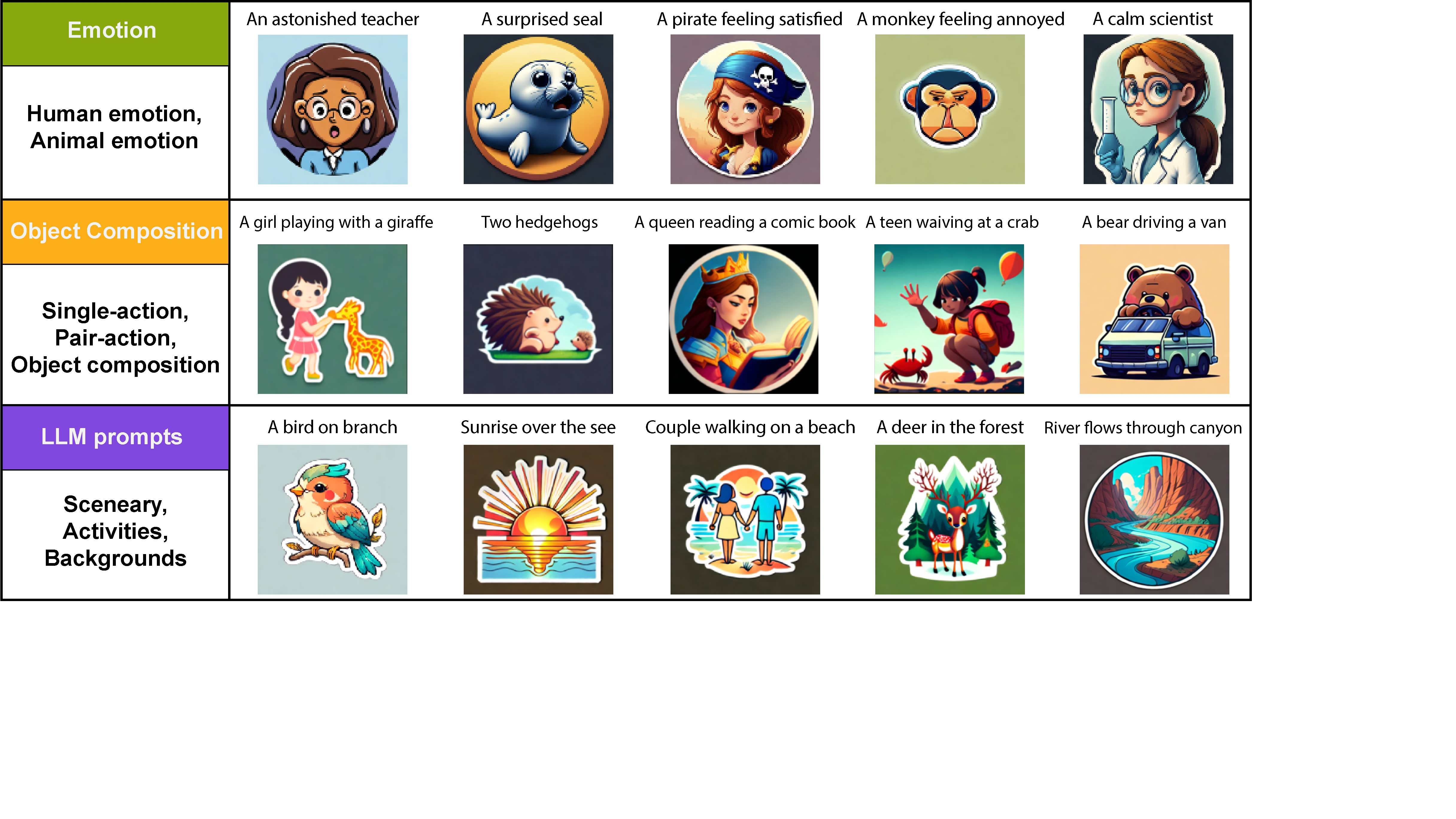}
    \caption{Stickers from each prompt bucket in the HITL Alignment dataset. All images are generated.}
    \label{fig:hitl_good_examples}
\end{figure}

As described in Section \ref{sec:datasets}, our HITL Alignment dataset contains prompts from three broad categories, i.e. (i) Emotion, (ii) Object Compositions, and (iii) Scene Diversity. We show some representative images from the HITL Alignment dataset in Fig. \ref{fig:hitl_good_examples}, and show the details on the pass-rate and number of training images in the HITL alignment dataset in Table \ref{tab:hitlprompts}.

\begin{table}[ht]\small
\caption{Summary of the HITL Alignment dataset. Images are generated from domain aligned model  and filtered by human annotators for good quality and high prompt alignment.}
\vspace{-15pt}
\begin{center}
    \resizebox{\linewidth}{!}{\begin{tabular}{c|cccc}
        \toprule
        \textbf{Prompt Bucket} & \textbf{Sub-category} & \textbf{\#Prompts} & \textbf{Pass-rate} & \textbf{\#Images}\\
        \midrule
        Emotion  & Human emotion & 2k & \multirow{2}{*}{0.383} &  \multirow{2}{*}{4.3k} \\  
        expressiveness & Animal emotion & 5k \\
        \hline
        Object  & Single action & 7.2k & \multirow{2}{*}{0.241} & \multirow{2}{*}{7.4k} \\
        composition & Pair action & 8.3k \\
        \hline
        Scene diversity & Scenes, activities, \textit{etc.} & 3.3k & 0.448 & 3k \\
        \bottomrule
    \end{tabular}}
\end{center}
\vspace{-8pt}
\label{tab:hitlprompts}
\end{table}

\subsection{EITL Style Dataset}
\begin{figure}[htbp]
    \centering
    \includegraphics[width=1\linewidth]{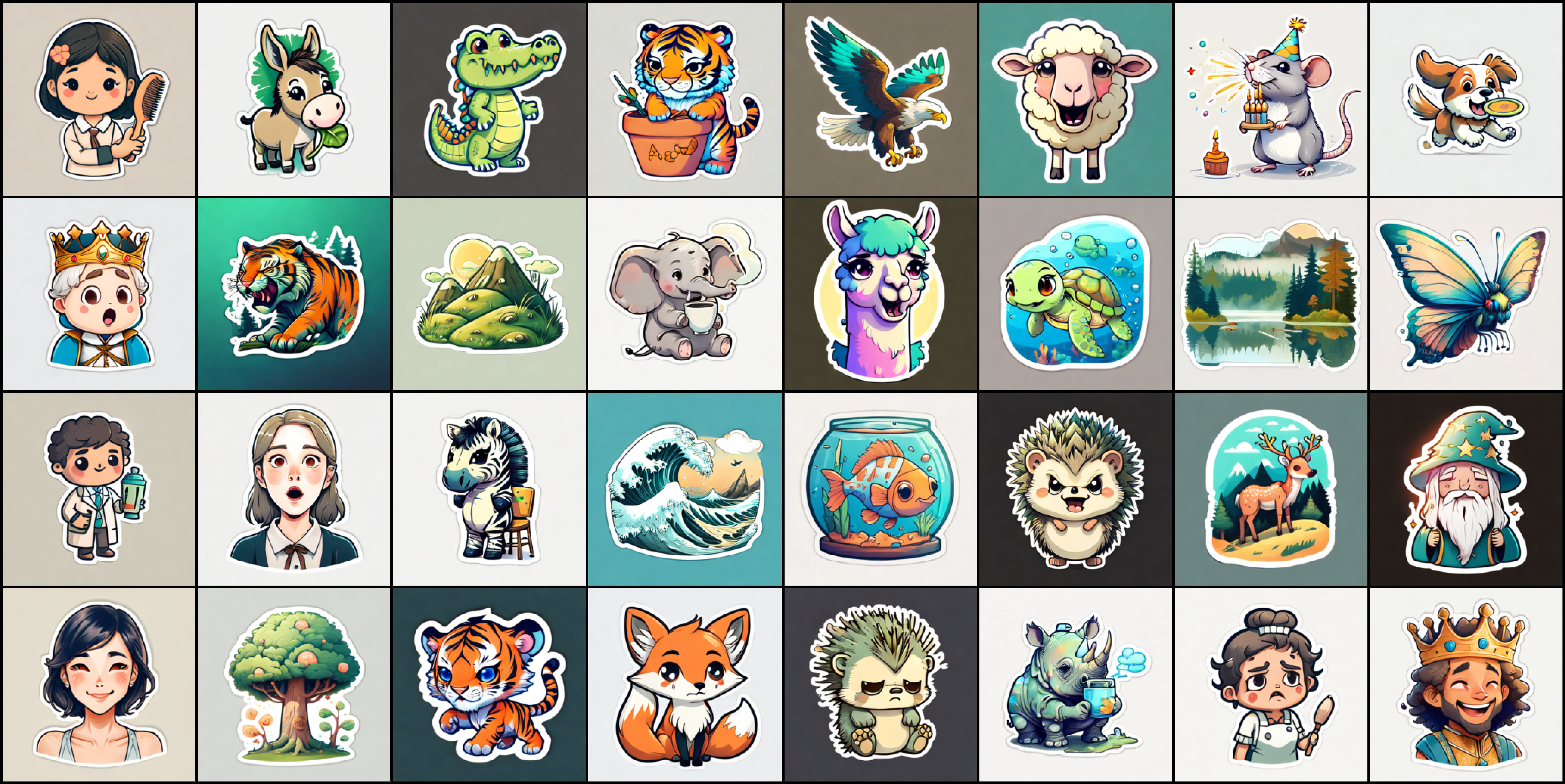}
    \caption{Stickers from the EITL Style dataset. All images are generated.}
    \label{fig:style_data_example}
\end{figure}

We collect our EITL Style dataset using Emu-256 with prompt engineering as described in Section \ref{sec:style_dataset} with the help of design experts.
Some sample stickers manually collected by the experts are shown in Fig. \ref{fig:style_data_example}.

\subsection{Visual Comparison of Finetuning Stages}
\begin{figure}[hbtp]
    \centering
    \includegraphics[width=1\linewidth]{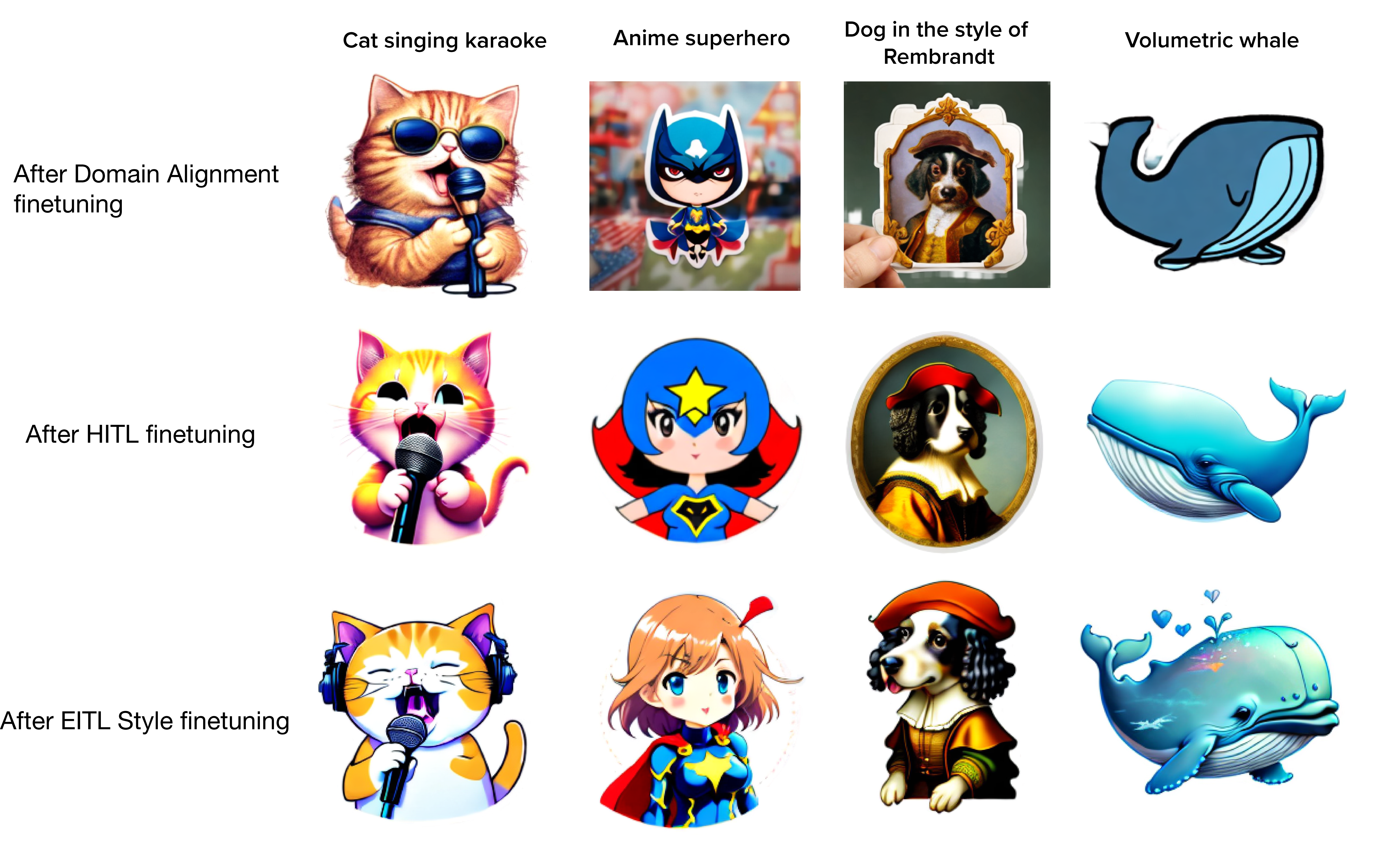}
    \caption{Results from each stage in standard finetuning to show the style evolution. Examples shown for each prompt are generated with the same starting noise. }
    \label{fig:effect_of_each_stage}
\end{figure}

In standard finetuning described in Section \ref{sec:alignmentfinetuning}, we observe that the style of generated stickers change after each finetuning step as we move closer to the target sticker style. We qualitatively show the effect of each finetuning stage in Fig. \ref{fig:effect_of_each_stage}. Following HITL finetuning, the model demonstrates improved semantic alignment with the sticker domain, while EITL Style finetuning enhances its visual quality and details. 

\begin{figure*}[t]
    \centering
    \includegraphics[width=1\linewidth]{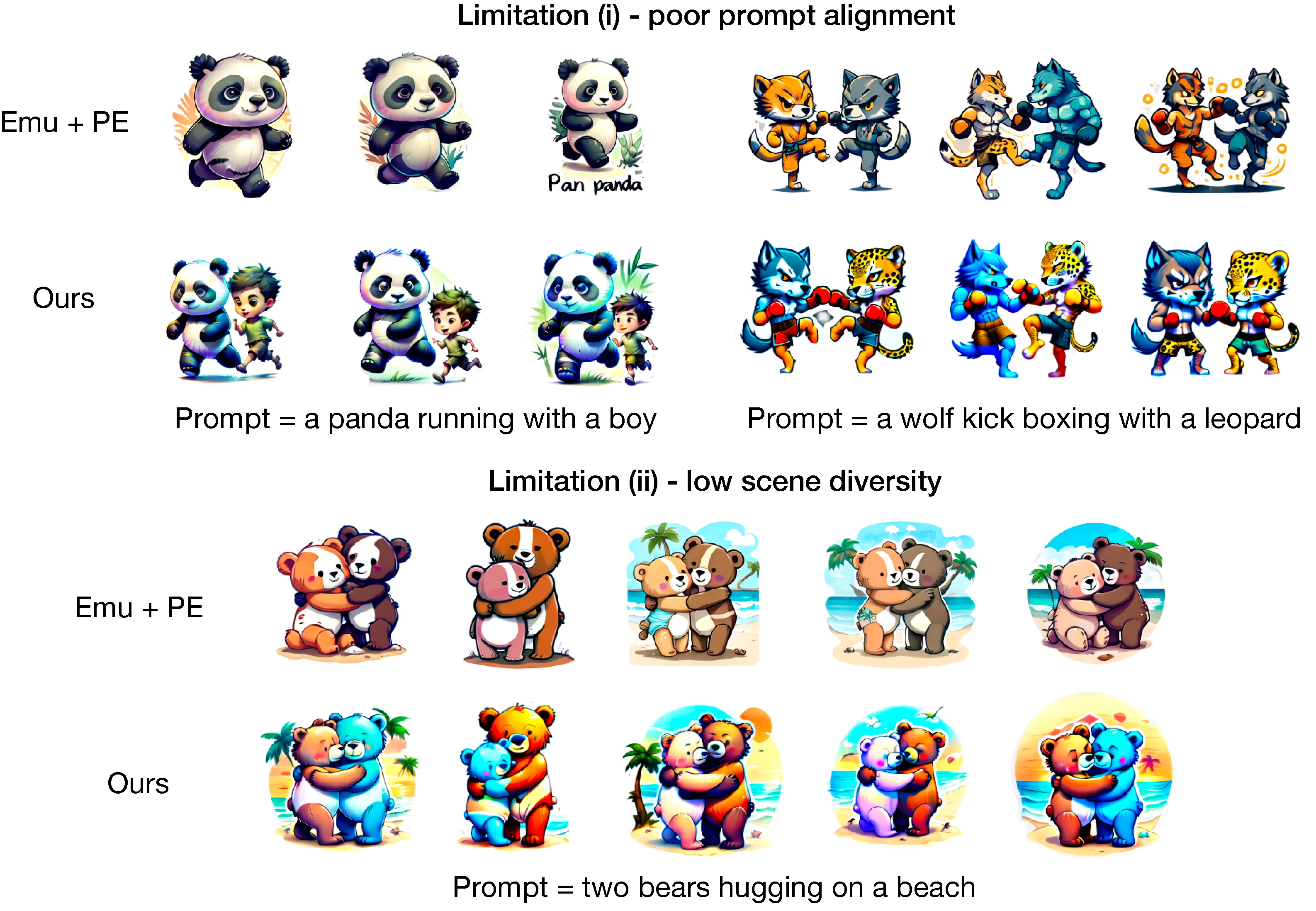}
    \caption{We observe two limitations when applying prompt engineering on the text-to-image model, Emu, to get the desired sticker style: (i) poor prompt alignment, where the model fails on object compositions, or conflates two entities into one, and (ii) low scene diversity, where the model always generates objects and scenes that are visually very similar. We address these limitations through our finetuning recipe and show improved results. We do not use the Prompt Enhancer module in this visualization.}
    \label{fig:limitation}
\end{figure*}

\section{Limitations of Emu-256 + PE baseline}
\label{sec:baseline_limitations}
As described in Section \ref{subsec:aba_baseline}, we apply prompt engineering (PE) on a strong text-to-image model, \textit{Emu} as the baseline. This baseline suffers from two limitations: (i) poor prompt alignment, and (ii) low scene diversity. We show some examples of the two limitations in Fig. \ref{fig:limitation}. In the top row, the model fails at generating the boy along with the panda, and leopard with the wolf, thus failing at generating images with multiple subjects. The model also conflates two entities into one, for instance, the leopard is generated with a wolf head in the second example in the first row. These two examples suggest poor prompt alignment in the baseline, where it either misses one entity completely, or conflates two entities into one. In the second row, the baseline generates objects that are visually very similar, for instance, the bears have similar color across generations and the background scene diversity is low as well. We address these two limitations using the HITL finetuning stage and show improved results.

\section{Additional Qualitative Results}
\label{sec:qualitative_results}
\begin{figure*}[ht]
    \centering
    \includegraphics[width=1\linewidth]{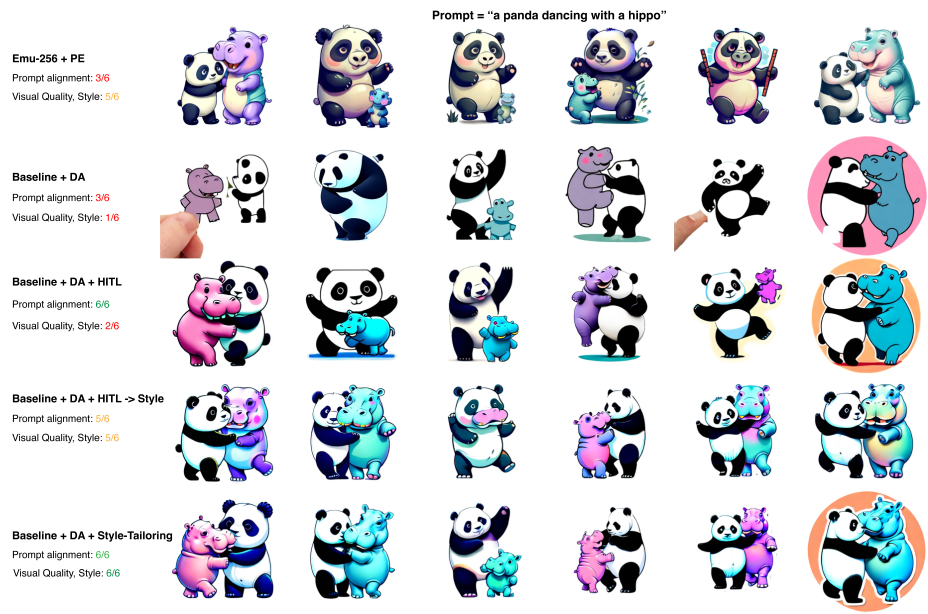}
    \caption{Qualitative comparison of the models with evaluation metrics shown in Table \ref{table:method_comparison}. Style Tailoring (Row 5) offers the best results in both prompt and style alignment.}
    \label{fig:more_examples_main_ablation}
\end{figure*}

In Fig. \ref{fig:more_examples_main_ablation}, we provide additional qualitative results of the models with evaluation metrics shown in Table \ref{table:method_comparison}.

\end{document}